\documentclass[10pt,twocolumn,letterpaper]{article}
\usepackage{multirow}
\usepackage[table]{xcolor}
\usepackage{cvpr}              
\definecolor{cvprblue}{rgb}{0.21,0.49,0.74}
\usepackage[pagebackref,breaklinks,colorlinks,allcolors=cvprblue]{hyperref}

\def\paperID{*****} 
\def\confName{CVPR}
\def\confYear{2026}

\title{Implicit-Scale 3D Reconstruction for Multi-Food Volume Estimation from Monocular Images}

\author{
Yuhao Chen\textsuperscript{1} \quad
Gautham Vinod\textsuperscript{2} \quad
Siddeshwar Raghavan\textsuperscript{2} \quad
Talha Ibn Mahmud\textsuperscript{2} \quad
Bruce Coburn\textsuperscript{2} \quad \\
Jinge Ma\textsuperscript{2} \quad 
Fengqing Zhu\textsuperscript{2} \quad
Jiangpeng He\textsuperscript{3}\(^{\dagger}\)
\\
\textsuperscript{1} University of Waterloo, Waterloo, Ontario, Canada \\
\textsuperscript{2} Purdue University, West Lafayette, Indiana, U.S.A. \\
\textsuperscript{3} Indiana University, Bloomington, Indiana, U.S.A. \\
}

\begin{document}
\maketitle

\renewcommand{\thefootnote}{\fnsymbol{footnote}} 
\footnotetext[2]{Corresponding author}

\begin{abstract}
We present \textit{Implicit-Scale 3D Reconstruction from Monocular Multi-Food Images}, a benchmark dataset designed to advance geometry-based food portion estimation in realistic dining scenarios. Existing dietary assessment methods largely rely on single-image analysis or appearance-based inference, including recent vision--language models, which lack explicit geometric reasoning and are sensitive to scale ambiguity. This benchmark reframes food portion estimation as an implicit-scale 3D reconstruction problem under monocular observations. To reflect real-world conditions, explicit physical references and metric annotations are removed; instead, contextual objects such as plates and utensils are provided, requiring algorithms to infer scale from implicit cues and prior knowledge. The dataset emphasizes multi-food scenes with diverse object geometries, frequent occlusions, and complex spatial arrangements. The benchmark was adopted as a challenge at the MetaFood 2025 Workshop, where multiple teams proposed reconstruction-based solutions. Experimental results show that while strong vision--language baselines achieve competitive performance, geometry-based reconstruction methods provide both improved accuracy and greater robustness, with the top-performing approach achieving \textbf{0.21 MAPE} in volume estimation and \textbf{5.7 L1 Chamfer Distance} in geometric accuracy.

\end{abstract}
\section{Introduction}
\label{sec:intro}
\begin{figure}[t]
    \centering
    \includegraphics[width=0.95\linewidth]{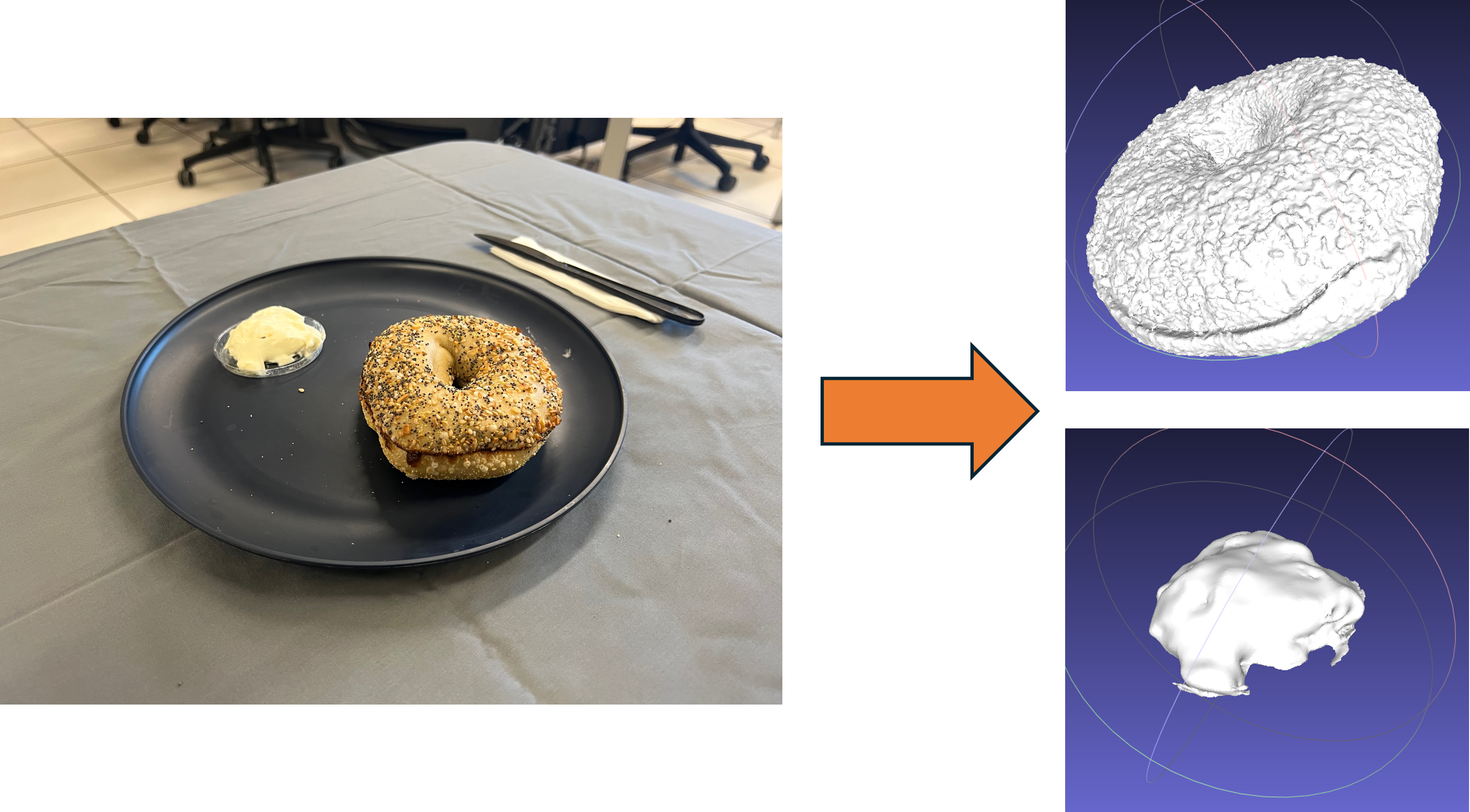}
    \vspace{-0.4cm}
    \caption{Implicit-scale 3D reconstruction of multi-food. Users are provided with a single image containing a realistic multi-food eating scenario, including multiple food items, plates and utensils with out explicit scale reference.}
    \label{fig:problem}
\end{figure}

3D object reconstruction has become an active research area in computer vision, driven by advances in multi-view geometry, neural rendering, and image-to-3D generation models \cite{neus, hunyuan3d}. By inferring object geometry from images, these methods enable estimation of physically meaningful quantities such as shape and volume, offering a principled alternative to appearance-based regression methods commonly used in food portion estimation.

Most existing dietary assessment approaches rely on single-image analysis, where food portions are estimated using learned visual features or segmentation-based heuristics \cite{konstantakopoulos_review_2024}. While effective in constrained settings, these methods lack explicit geometric reasoning and often struggle with generalization across food categories, presentations, and viewpoints. In contrast, 3D reconstruction-based approaches provide a direct mechanism for estimating food volume by grounding predictions in reconstructed geometry rather than image statistics alone.

Despite their promise, current 3D reconstruction paradigms face important limitations in real-world eating scenarios. Multi-view reconstruction methods \cite{neus} require users to capture multiple images from different viewpoints, introducing interaction overhead that is impractical for everyday dietary tracking. Recent single-image image-to-3D generation techniques \cite{hunyuan3d} reduce this burden but typically assume isolated objects and controlled environments. As a result, these methods do not generalize well to realistic dining scenes, where multiple food items are simultaneously present.

Reconstructing multiple food objects from a single image is particularly challenging due to frequent occlusions, overlapping geometries, shadows, and complex spatial arrangements. Furthermore, food exhibits high intra-class variability: the same dish can appear in diverse shapes, sizes, and layouts across instances. These properties limit the effectiveness of approaches based on predefined templates, canonical shapes, or fixed layouts. In addition, most monocular 3D reconstruction methods suffer from scale ambiguity~\cite{vinod2026size, vinod2024food, ma2024mfp3d}, while image-to-3D generation models  \cite{hunyuan3d} commonly normalize object scale and pose during training, resulting in reconstructed geometries without consistent metric interpretation across scenes.

In this work, we focus on implicit-scale 3D reconstruction in multi-food scenarios, reflecting realistic eating occasions where multiple items with diverse geometries and material properties coexist on a plate. Instead of providing explicit physical references \cite{he2024metafoodcvpr2024challenge} or metric annotations, we intentionally remove such cues and require algorithms to infer scale from implicit contextual information, such as plates, utensils, and commonly occurring food items with familiar size priors. This formulation reframes food portion estimation as a physics-informed 3D reconstruction problem under monocular scale ambiguity, as shown in Fig. \ref{fig:problem}. 

To support this direction, we introduce \textit{Implicit-Scale 3D Reconstruction from Monocular Multi-Food Images}, a benchmark dataset designed to evaluate reconstruction-based food volume estimation under realistic dining conditions. The benchmark emphasizes multi-food scenes with diverse object appearances, frequent occlusions, and ambiguous scale cues, reflecting challenges commonly encountered in real-world eating scenarios. It was adopted as a challenge at the MetaFood 2025 Workshop \cite{kaggle_multifood3drecon}, where multiple teams proposed monocular reconstruction-based solutions. In addition to reconstruction pipelines, we evaluate a strong vision--language model baseline for volume estimation. While such appearance-based methods can perform competitively, benchmark results demonstrate that approaches explicitly modeling 3D geometry and implicit scale provide both higher accuracy and improved robustness across food categories. In particular, the first-place solution illustrates how joint reconstruction and scale reasoning lead to stable performance in complex multi-food scenes.

\begin{figure}[t]
    \centering
    \includegraphics[width=0.95\linewidth]{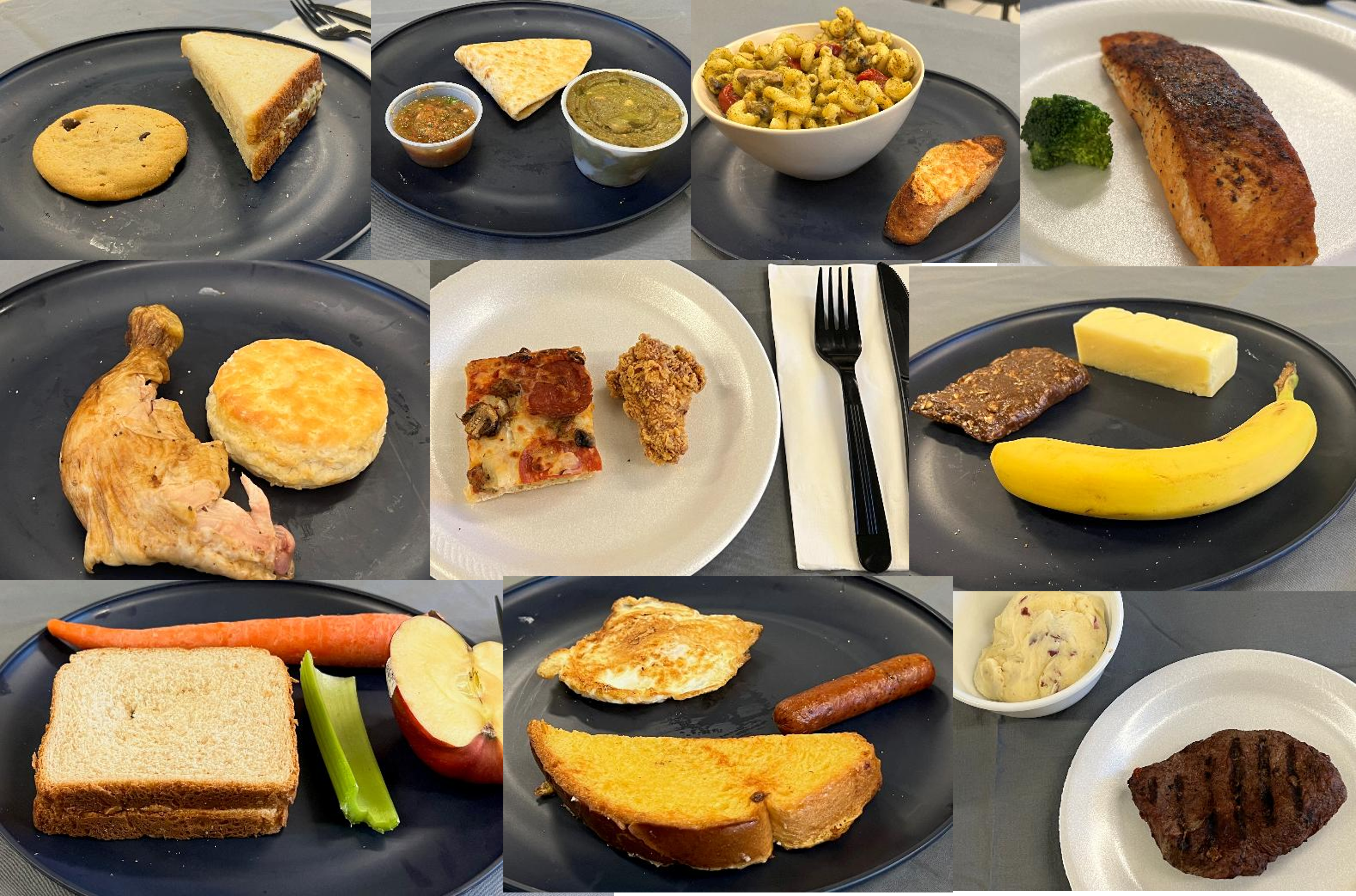}
    \vspace{-0.2cm}
\caption{Cropped examples from our benchmark dataset. Original images include a wider field of view containing utensils and surrounding context.}

    \label{fig:benchmark}
\end{figure}

\vspace{-0.3cm}
\section{Dataset}
\vspace{-0.3cm}



Our benchmark dataset comprises 10 curated multi-food scenes, each containing multiple food items accompanied by utensils and plates, for a total of 24 3D objects, aligning with the MetaFood 2025 Workshop challenge. The food combinations are selected from the MetaFood3D~\cite{chen2024metafood3d} object set to reflect realistic eating scenarios. All items are captured using a high-precision 3D scanner following the MetaFood3D data collection pipeline. Utensils and plates are intentionally included as implicit physical references for scale, enabling evaluation of physically accurate reconstruction and volume estimation under monocular scale ambiguity. Each food item is individually scanned and annotated, while scene difficulty varies with the number of food items and the degree of occlusion induced by camera viewpoints. 
Fig. \ref{fig:benchmark} shows cropped examples of our benchmark.

\section{Methodologies}
\label{sec:method}
\begin{table*}[t]
\centering
\caption{Three representative classes of methods}
\label{tab:recon_scaling}
\small
\begin{tabular}{l l l l}
\hline
\textbf{Method} & \textbf{Reconstruction Strategy} & \textbf{Scaling Mechanism} & \textbf{Key Reference} \\
\hline
\multirow{2}{*}{PSHS}
 & \multirow{2}{*}{Independent (per-object)}
 & \textbf{Pixel-space heuristic}: ratio of object-to-plate
 & \multirow{2}{*}{Image-space plate bounding box} \\
 &  & bounding box diagonals (single reference) & \\

\multirow{2}{*}{SGPS}
 & \multirow{2}{*}{Independent (per-object)}
 & \textbf{Scene-level geometric prior}: uniform scale
 & \multirow{2}{*}{Known plate / utensil dimensions} \\
 &  & derived from reconstructed plate and utensil meshes & \\

\multirow{3}{*}{MDMS}
 & \multirow{3}{*}{Joint (scene-level)}
 & \textbf{Metric Depth Driven Multi-stage scaling}:
 & \multirow{3}{*}{Metric depth + web-crawled priors} \\
 &  & (1) local 6D pose alignment & \\
 &  & (2) global scale refinement using web data & \\

\hline
\end{tabular}
\end{table*}

This section analyzes three representative classes of methods for implicit scale estimation in monocular multi-food 3D reconstruction, as exemplified by the top-performing submissions in the MetaFood3D 2025 Challenge. We categorize the approaches based on how scale is inferred and propagated across reconstructed objects.

All methods adopt Hunyuan3D \cite{hunyuan3d} as the shared monocular image-to-3D reconstruction backbone. The primary distinction lies in how physical scale is introduced, spanning pixel-space heuristics, scene-level geometric priors, and depth-driven multi-stage scaling pipelines. Table~\ref{tab:recon_scaling} summarizes the key characteristics of each category.

\subsection{Pixel-Space Heuristic Scaling (PSHS)}
Methods \cite{MetaFood2025_Challenge1} in this category reconstruct each food object independently. Object instances are segmented in the image and individually reconstructed into separate 3D meshes. 
Scale estimation is performed by detecting a plate in the image and measuring its diagonal length in pixel space. A fixed physical plate size is assumed, and a scale factor is computed as the ratio between the assumed real-world plate diagonal and its pixel-space measurement. This scale factor is then applied to each reconstructed food object by scaling the extremal points of its corresponding 3D mesh.

\subsection{Scene-Level Geometric Prior Scaling (SGPS)}
This category \cite{MetaFood2025_Challenge1} reconstructs all objects in the scene individually, including food items, plates, and utensils.
Predefined physical dimensions of plates and utensils are used as geometric priors to compute a single global scale factor. This scale factor is applied uniformly to all reconstructed objects, aligning the reconstructed scene with real-world dimensions and enabling volume estimation from the scaled meshes.

\subsection{Metric Depth–Driven Multi-Stage Scaling (MDMS)}
This category \cite{monobite} integrates joint reconstruction, metric depth estimation, and multi-stage scale refinement. All objects in the scene are first reconstructed as a single combined mesh, enforcing consistent relative scale and spatial alignment across objects.

Individual objects are subsequently separated in 3D space using k-means clustering, producing per-object meshes that share a common scale reference. Metric scale is estimated using monocular depth prediction \cite{depthpro}, with local scale obtained by aligning the reconstructed mesh to a metric-depth point cloud through a coarse-to-fine 6D pose estimation pipeline. This alignment optimizes a similarity transform between mesh surface points and depth observations.

To address global scale ambiguity in monocular depth estimation, external size priors obtained from web-crawled statistics of common plates and utensils are used to refine the scale. The final object scale is computed as the product of the local depth-derived scale and a global correction factor derived from these priors.

\section{EXPERIMENTS AND DISCUSSION}

\begin{table}[t]
\centering
\caption{Per-food error comparison (lower is better). Best and second-best are highlighted in green and yellow, respectively.}
\label{tab:per_food_errors}
\footnotesize
\setlength{\tabcolsep}{4pt}
\begin{tabular}{lcccc}
\hline
\textbf{Food} & \textbf{GPT 5.2} & \textbf{PSHS} & \textbf{SGPS} & \textbf{MDMS} \\
\hline
energy\_bar      & 0.48 & 0.55 & \cellcolor{yellow!20}0.44 & \cellcolor{green!20}0.21 \\
cheddar\_cheese  & 0.48 & 0.60 & \cellcolor{green!20}0.07 & \cellcolor{yellow!20}0.22 \\
banana           & 0.29 & 0.35 & \cellcolor{yellow!20}0.15 & \cellcolor{green!20}0.09 \\
grilled\_salmon  & 0.43 & \cellcolor{green!20}0.32 & \cellcolor{yellow!20}0.35 & 0.46 \\
pasta            & 0.52 & 0.75 & \cellcolor{yellow!20}0.12 & \cellcolor{green!20}0.11 \\
garlic\_bread    & 0.45 & \cellcolor{yellow!20}0.35 & 0.40 & \cellcolor{green!20}0.23 \\
pb\&j            & 0.43 & \cellcolor{yellow!20}0.22 & 0.55 & \cellcolor{green!20}0.01 \\
carrot\_stick    & 0.77 & 0.95 & \cellcolor{yellow!20}0.40 & \cellcolor{green!20}0.04 \\
apple            & 0.32 & 0.50 & \cellcolor{green!20}0.09 & \cellcolor{yellow!20}0.28 \\
celery           & 0.32 & \cellcolor{green!20}0.16 & 1.08 & \cellcolor{yellow!20}0.17 \\
pizza            & \cellcolor{yellow!20}0.05 & 0.80 & \cellcolor{green!20}0.03 & 0.30 \\

chicken\_wing    & \cellcolor{yellow!20}0.17 & 0.21 & \cellcolor{green!20}0.09 & 0.21 \\
quesadilla       & \cellcolor{green!20}0.22 & 0.67 & 0.69 & \cellcolor{yellow!20}0.66 \\
guacamole        & \cellcolor{green!20}0.11 & 0.72 & 0.46 & \cellcolor{yellow!20}0.32 \\

salsa            & \cellcolor{green!20}0.21 & 0.90 & \cellcolor{yellow!20}0.32 & 0.51 \\
roast\_chicken\_leg & \cellcolor{yellow!20}0.12 & \cellcolor{green!20}0.04 & 0.50 & 0.22 \\
biscuit          & 0.44 & \cellcolor{yellow!20}0.12 & \cellcolor{green!20}0.04 & 0.29 \\
sandwich         & 0.60 & 0.58 & \cellcolor{yellow!20}0.12 & \cellcolor{green!20}0.11 \\
cookie           & 0.44 & 0.49 & \cellcolor{green!20}0.13 & \cellcolor{yellow!20}0.29 \\
steak            & 0.19 & \cellcolor{yellow!20}0.16 & 0.22 & \cellcolor{green!20}0.04 \\

mashed\_potatoes & \cellcolor{yellow!20}0.35 & 0.36 & 0.58 & \cellcolor{green!20}0.05 \\

toast            & 0.53 & 0.25 & \cellcolor{yellow!20}0.07 & \cellcolor{green!20}0.01 \\
sausage          & 0.38 & 0.52 & \cellcolor{green!20}0.21 & \cellcolor{yellow!20}0.23 \\
fried\_egg       & \cellcolor{green!20}0.03 & 0.63 & 0.48 & \cellcolor{yellow!20}0.08 \\
\hline

MAPE             & 0.34 & 0.46 & \cellcolor{yellow!20}0.31 & \cellcolor{green!20}0.21 \\
STD             & \cellcolor{yellow!20}0.18 & 0.25 & 0.25 & \cellcolor{green!20}0.16\\
\hline
\end{tabular}
\end{table}

\begin{table}[t]
\centering
\caption{Per-food L1 Chamfer distance comparison (lower is better). Best and second-best are highlighted in green and yellow, respectively.}
\label{tab:l1_chamfer}
\footnotesize
\setlength{\tabcolsep}{5pt}
\begin{tabular}{lccc}
\hline
\textbf{Food} & \textbf{PSHS} & \textbf{SGPS} & \textbf{MDMS} \\
\hline
energy\_bar        & 9.15 & \cellcolor{yellow!20}4.92 & \cellcolor{green!20}2.78 \\
cheddar\_cheese    & 12.65 & \cellcolor{green!20}3.33 & \cellcolor{yellow!20}4.44 \\
banana             & 9.97 & \cellcolor{green!20}5.67 & \cellcolor{yellow!20}7.01 \\
grilled\_salmon    & 13.67 & \cellcolor{green!20}12.48 & \cellcolor{yellow!20}12.88 \\
pasta              & 29.13 & \cellcolor{yellow!20}6.88 & \cellcolor{green!20}6.35 \\
garlic\_bread      & 7.36 & \cellcolor{green!20}4.62 & \cellcolor{yellow!20}6.44 \\
pb\&j              & 5.81 & 7.35 & \cellcolor{green!20}4.58 \\
carrot\_stick      & 15.39 & \cellcolor{yellow!20}4.50 & \cellcolor{green!20}3.69 \\
apple              & \cellcolor{green!20}8.87 & 9.26 & \cellcolor{yellow!20}9.09 \\
celery             & 12.65 & \cellcolor{yellow!20}7.26 & \cellcolor{green!20}3.18 \\
pizza              & 21.70 & \cellcolor{green!20}3.43 & \cellcolor{yellow!20}3.75 \\
chicken\_wing      & \cellcolor{yellow!20}3.90 & \cellcolor{green!20}3.80 & 4.85 \\
quesadilla         & 13.75 & \cellcolor{yellow!20}9.31 & \cellcolor{green!20}8.16 \\
guacamole          & 9.78 & \cellcolor{yellow!20}4.62 & \cellcolor{green!20}2.48 \\
salsa              & 10.25 & \cellcolor{yellow!20}6.68 & \cellcolor{green!20}3.40 \\
roast\_chicken\_leg& 13.05 & \cellcolor{yellow!20}7.96 & \cellcolor{green!20}6.56 \\
biscuit            & \cellcolor{green!20}2.61 & \cellcolor{yellow!20}5.24 & 2.83 \\
sandwich           & 15.39 & \cellcolor{green!20}6.80 & \cellcolor{yellow!20}10.46 \\
cookie             & 24.65 & \cellcolor{yellow!20}4.65 & \cellcolor{green!20}3.83 \\
steak              & \cellcolor{green!20}5.33 & 17.10 & \cellcolor{yellow!20}8.42 \\
mashed\_potatoes   & 17.22 & \cellcolor{yellow!20}16.97 & \cellcolor{green!20}6.46 \\
toast              & \cellcolor{green!20}6.39 & 9.49 & \cellcolor{yellow!20}6.58 \\
sausage            & 7.24 & \cellcolor{green!20}3.18 & \cellcolor{yellow!20}3.55 \\
fried\_egg         & 9.45 & \cellcolor{green!20}4.24 & \cellcolor{yellow!20}5.05 \\
\hline
Average   & 11.89 & \cellcolor{yellow!20}7.07 & \cellcolor{green!20}5.70 \\
STD                & 6.44  & \cellcolor{yellow!20}3.85 & \cellcolor{green!20}2.66 \\
\hline
\end{tabular}
\end{table}

\textbf{Experimental Setup.} 
We evaluate the three method categories (PSHS, SGPS, and MDMS) using Mean Absolute Percentage Error (MAPE) for food volume estimation and L1 Chamfer Distance for geometric reconstruction accuracy. Lower values indicate better performance for both metrics. In addition, we include GPT-5.2 \cite{openai_gpt52} as a reference baseline for volume estimation, representing a vision–language model that infers food volume without explicit 3D reconstruction.

To avoid errors from automated 6D pose estimation, reconstructed 3D objects are manually transformed by each team to align with the reference ground-truth point clouds prior to Chamfer distance computation. This alignment ensures that geometric error reflects reconstruction quality rather than pose misalignment. Volume estimation is evaluated directly from the aligned meshes using MAPE.

\textbf{Results and Discussion.}
As shown in Table~\ref{tab:per_food_errors}, the metric-depth driven method MDMS achieves the lowest overall volume error with a MAPE of 0.21, outperforming the scene geometry prior method SGPS (0.31), pixel-space hueristic method PSHS (0.46), and GPT-5.2 (0.34). It also exhibits the lowest variance across food categories, indicating more stable performance under diverse multi-food scene configurations. SGPS provides a clear improvement over PSHS, suggesting that leveraging multiple reconstructed reference objects yields more reliable implicit scale estimation than single-reference heuristics.

Geometric reconstruction results in Table~\ref{tab:l1_chamfer} follow a consistent trend. MDMS achieves the lowest average L1 Chamfer Distance (5.70), followed by SGPS (7.07) and PSHS (11.89) methods, with correspondingly lower variance. This alignment between geometric accuracy and volume estimation performance highlights the importance of explicit 3D reconstruction for robust food portion estimation.

GPT-5.2 performs competitively on volume estimation and notably outperforms the pixel-space heuristic method PSHS, whose pixel-based scaling relies on bounding-box measurements that are sensitive to bounding box noise and object orientation. However, GPT-5.2 does not produce explicit 3D geometry or pose estimates, preventing direct geometric evaluation and limiting its applicability to downstream geometry-aware tasks.

Notably, although 6D pose alignment is manually performed for evaluation across all submissions, MDMS explicitly incorporates a coarse-to-fine 6D pose estimation pipeline as part of its reconstruction process. While not required for benchmark evaluation, this capability provides additional information that may benefit downstream applications such as robotic grasping, manipulation, and scene understanding.

Overall, the results demonstrate that methods explicitly modeling 3D geometry and implicit scale achieve lower error and greater robustness than appearance-based inference.

\section{Conclusion}
We introduced a benchmark for implicit-scale 3D reconstruction from monocular multi-food images under realistic dining conditions. Experimental results show that while strong vision--language baselines achieve competitive volume estimates, geometry-based reconstruction methods provide both higher accuracy and greater robustness, as demonstrated by the top-performing solution. We hope this benchmark will encourage further research on geometry-aware reconstruction for reliable food portion estimation.

{
    \small
    \bibliographystyle{ieeenat_fullname}
    \bibliography{main}
}
\end{document}